%% file: main.tex
\documentclass[letterpaper, 10 pt, conference]{ieeeconf}
\IEEEoverridecommandlockouts                              
\input{preamble.tex}
\usepackage{censor}
\usepackage{todonotes}
\usepackage{siunitx}

\begin{document}


\title{
 \LARGE \bf ContactFusion: Stochastic Poisson Surface Maps\\ from Visual and Contact Sensing
}

\author{
Aditya Kamireddypalli$^1$, Jo\~{a}o Moura$^1$, Russell Buchanan$^2$, Mat{\'i}as Mattamala$^{1}$, \\Sethu Vijayakumar$^1$, Subramanian Ramamoorthy$^1$
%
\thanks{$^1$ School of Informatics, University of Edinburgh, Scotland. $\texttt{a.kamireddypalli@sms.ed.ac.uk}$}
\thanks{$^2$ Department of Mechanical and Mechatronics Engineering, University of Waterloo, Canada.}
\thanks{\noindent A. Kamireddypalli is supported in part by funding in the form of an unrestricted gift from Microsoft Research. S. Ramamoorthy is supported by a UKRI Turing AI World Leading Researcher Fellowship on AI for Person-Centred and Teachable Autonomy (grant EP/Z534833/1).}
\thanks{\noindent For the purpose of open access, the authors have applied a Creative Commons Attribution (CC BY) license to any Author Accepted Manuscript version arising from this submission.}
}

\maketitle


\begin{abstract}
Robotic assembly tasks such as peg-in-hole insertion require precise geometric reasoning, yet sensor noise in real-world systems often exceeds the tight tolerances required for successful insertion.
Vision-based pose estimation alone can therefore lead to misalignment and failure.
In this work, we propose \textit{ContactFusion}, a probabilistic mapping framework that fuses depth sensing and force–torque measurements to estimate the geometry of insertion targets.
Our method builds a Stochastic Poisson Surface Map (SPSMap), an uncertainty-aware implicit surface representation constructed using Stochastic Poisson Surface Reconstruction (SPSR).
To incorporate contact information, we introduce a contact location estimator that converts force–torque measurements into spatial hypotheses over candidate contact locations on the robot end-effector.
These hypotheses are fused with depth observations within a sequential reconstruction framework, enabling the map to be refined through both visual and contact interactions.
We evaluate ContactFusion in simulation and on a real robotic system in a peg-in-hole setting.
Our results show that SPSMap produces more accurate and geometrically consistent reconstructions, improving reconstruction F-score by up to 30--35\%, while providing uncertainty estimates that enable active reconstruction strategies.
\end{abstract}

\input{content/intro_revised}

\input{content/related_revised}

\input{content/background_revised}
\input{content/method_revised_v2}
\input{content/evaluation_revised}

\input{content/results}


\input{content/conclusion}
\printbibliography 

\end{document}

%% file: preamble.tex
\usepackage[
    backend=bibtex,
    style=ieee,
    mincitenames=1,
    maxcitenames=2,
    maxbibnames=20,
    isbn=false,
    url=false,
    doi=false,
    natbib=true]{biblatex}
\let\labelindent\relax 
\usepackage{graphicx} 
\usepackage{times}
\usepackage{soul}
\usepackage{xcolor}
\usepackage{multirow}
\usepackage{booktabs}
\usepackage{amsmath}
\usepackage{amssymb}
\usepackage{subcaption}
\usepackage{algorithm}
\usepackage{algpseudocode}
\usepackage{svg}
\usepackage{enumitem}

\usepackage{tikz}

\usepackage{multicol}
\usepackage[bookmarks=true]{hyperref}
\usepackage{doi} 
\usepackage[capitalise,nameinlink]{cleveref} 

\def\secref#1{Sec.~\ref{#1}}

\def\figref#1{Fig.~\ref{#1}}

\def\eqref#1{Eq.~(\ref{#1})}

%% file: content/intro_revised.tex
\section{Introduction}
Dexterous manipulation often requires precise geometric reasoning; however, sensor noise in real robot applications often exceeds the necessary tolerances.
Tasks such as part alignment, constrained sliding, snap-fitting, and peg-in-hole insertion demand accurate estimates of the relative pose between interacting objects.
As tolerances tighten, millimeter-scale geometric errors can lead to jamming, excessive contact forces, or failure to complete the task. This makes geometric state estimation a crucial component for the success of current manipulation systems.

Current manipulation pipelines determine the pose of the insertion target by geometric registration of RGB-D measurements against a reference model (e.g., CAD model)~\cite{haralick_pose_1989}, or using deep learning-based prediction methods~\cite{wen_foundationpose_2024}.
While vision-based solutions can estimate the target's pose up to the millimeter-level, they can be affected by sensor noise, environmental and robot-induced occlusions, producing misalignment. Exploiting passive mechanical compliance and robust control strategies, such as impedance or compliant control, can reduce contact forces and compensate for this misalignment, at the cost of introducing a tradeoff between compliance and trajectory tracking \cite{jiang_review_2022}. One way to tackle such a requirement for precision could be to complement global visual information with fine local contact information.
Motivated by recent works \cite{kim_active_2022}, we explore this problem from the perspective of building multi-modal scene representations.

\begin{figure}[t!]
    \centering
    \includegraphics[width=\linewidth]{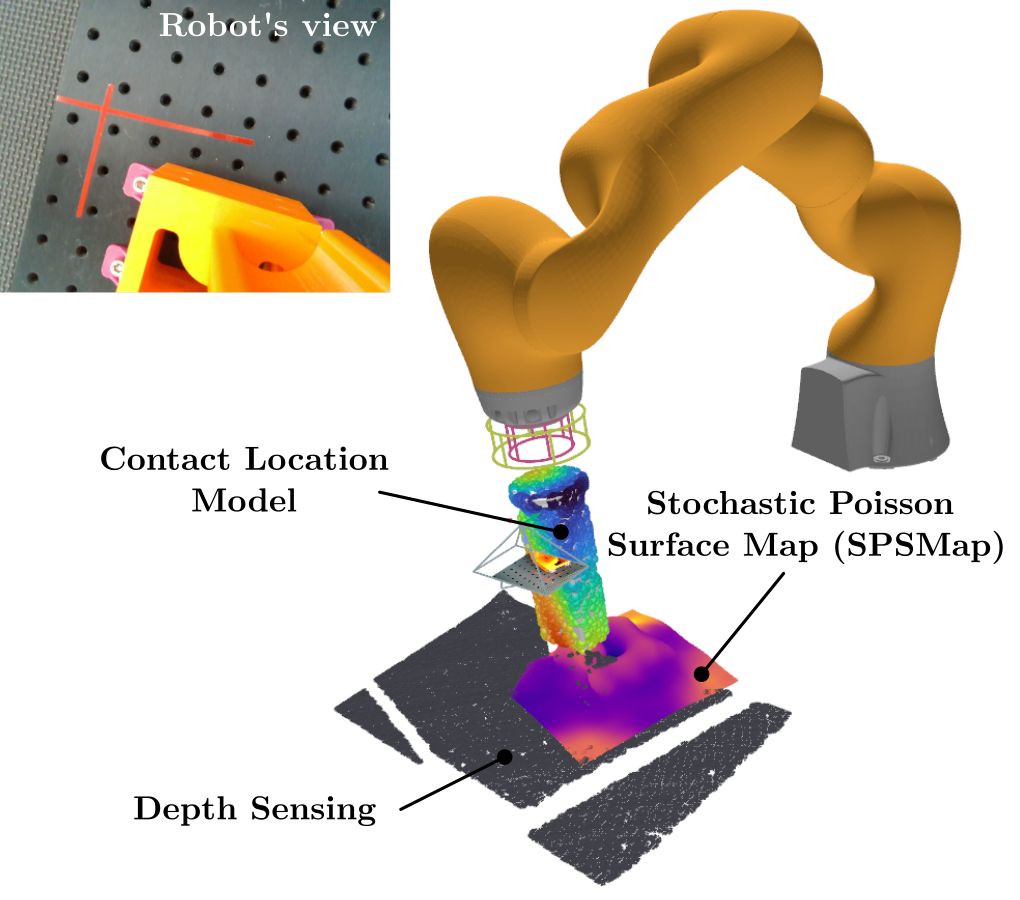}
\caption{
\textbf{ContactFusion builds probabilistic maps of an insertion target.} By building upon the Stochastic Poisson Surface Reconstruction method \cite{sellan_stochastic_2022} in a sequential formulation and a new contact location model fused with depth observations, ContactFusion produces continuous estimates of the target's surface and uncertainty that are up to 30\% more accurate than prior approaches.
}
\label{fig:motivation}
\end{figure}

In this work, we introduce \textit{ContactFusion}, a probabilistic mapping framework that builds a geometric representation of the insertion target by fusing depth and contact measurements.
Our framework integrates these different sensor modalities using the Stochastic Poisson Surface Reconstruction (SPSR) method~\cite{sellan_stochastic_2022}, producing maps that explicitly model the surface of the target object and its uncertainty.
We experimentally demonstrate that by combining vision and contact into this probabilistic representation, we can improve estimates of the target's shape by 30\% compared to existing methods, uncertainty estimates of the geometry, and also enable further directions such as active perception.


Our specific contributions are:
\begin{itemize}
    \item We present a probabilistic mapping framework which combines vision and contact sensing using the Stochastic Poisson Surface Reconstruction method, into an uncertainty-aware representation---\textit{SPSMap}.
    \item We introduce a contact location estimator to transform force/torque measurements into spatial likelihoods over candidate surface contact locations.
    \item We validate our framework through simulation and real-world experiments, demonstrating up to 30\% improvements in 3D reconstruction accuracy of the insertion target with respect to alternative mapping baselines.
    \item We further demonstrate how SPSMap enables uncertainty-driven active perception, accelerating the identification of the insertion target.
\end{itemize}

%% file: content/related_revised.tex
\section{Related Work}


\subsection{Manipulation-enhanced Mapping}
We first focus on prior mapping methods which are explicitly focused on the manipulation setting.
Interactive Perception~\cite{bohg_interactive_2016} is a general approach whereby an agent gains information about its environment through continuous perception and interaction while solving a task. This makes it ideal for insertion tasks, where the robot continuously perceives the insertion target and interacts with it. Inspired by this approach, we focus on the problem of representing the insertion target from vision and contact sensing.

The work by \citet{dragiev_gaussian_2011} is one of the earliest works that merged visual and tactile sensing into a common scene representation. Building upon a Gaussian Process Implicit Surfaces (GPIS)~\cite{williams2007gaussian} representation, they were able to combine haptic and 3D sensing while explicitly modelling uncertainty in the scene reconstruction. The approach was later extended to online estimation of 3D surfaces, demonstrating it in robotic grasping tasks \cite{caccamo_active_2016}, as well as for estimation of surface and object pose from tactile sensing \cite{suresh_tactile_2021}.

Occupancy maps are a well-established map representation, that has also been used for fusing contact and visual sensing. \citet{murali_shared_2024} used an occupancy map to to infer geometry and pose of objects using tactile and depth sensing, achieving improved manipulation efficiency within the context of an active perception pipeline~\cite{murali_shared_2024}. The occupancy probability stored in occupancy maps has also been used to guide non-prehensile maneuvers in kinematically-constrained spaces using information gain strategies~\cite{dengler_viewpoint_2023, marques_map_2025}.

More recently, neural implicit representations have also opened new avenues for joint object surface reconstruction and pose estimation for in-hand object manipulation~\cite{suresh_neural_2023}. However, these approaches are data intensive, requiring an offline training stage, and do not encode reconstruction uncertainty as part of the same model.

In this work, we explicitly aim to build a representation from online streams, fusing contact and vision sensing into a common representation for surface's geometry and uncertainty. To achieve this, we build upon the GPIS paradigm and extend it using the Stochastic Poisson Surface Reconstruction (SPSR)~\cite{sellan_stochastic_2022}. This enables us not only to fuse different sensing modalities, but also to perform statistical queries like on-surface probability and occupancy probability.


\subsection{Contact Location Estimation}
While the previous approaches have fused contact sensing with vision, they assume that the contact location is known. Previous works have proposed different strategies to determine it, depending on the sensor modalities available in each specific robot setup.

When force/torque (F/T) sensors are available, it has been proposed to use precomputed force-torque maps in Euclidean space \cite{ForceTorqueMap}. These maps relate expected force-torque values to Euclidean increments in robot end-effector positions . This map is computed offline using CAD information, with force-torque sensory information being used to query the map online. \citet{localisationAssembly} extended this idea by constructing a map using force-torque information in configuration space.

When no F/T sensors are available, the robot kinematics model can be used as a proxy for location, up to dynamic effects. \citet{8793572} followed this approach to implement a state estimation pipeline to determine the location of a known manipulation target in contact-rich scenarios.

With the development of vision-based tactile sensors, it has been possible to detect contact with higher fidelity. \citet{kim_active_2022} used a GelSight sensor placed at the fingertips of a robot gripper to locate the edges of the manipulation object. This additional information was shown to improve the performance of the manipulation policy in insertion tasks.

In ContactFusion we introduce a \textit{contact location model}, which relies on F/T sensing to enable the fusion of vision and contact measurements into a unified representation. This model provides a contact likelihood that allows us to handle the different noise models, and enable probabilistic fusion of the measurements, consequently improving the uncertainty estimates of the map.

%% file: content/background_revised.tex
\section{Background}
Before presenting our ContactFusion method, we introduce the basic algorithms it builds upon.

\begin{figure}[t]
    \centering
    \includegraphics[width=1\linewidth]{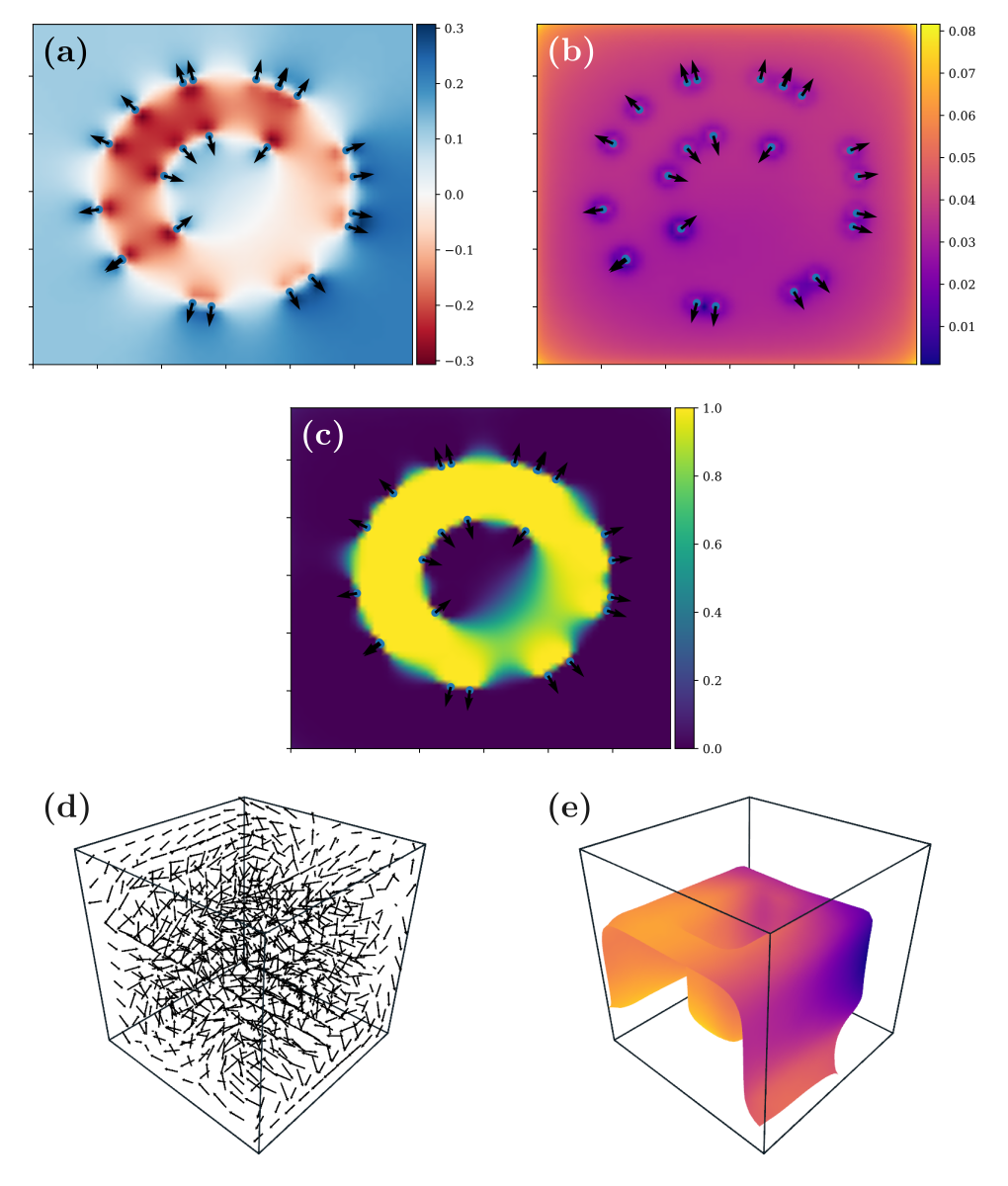}
    \caption{\textbf{Illustrative example of the Stochastic Poisson Surface Reconstruction (SPSR).} (a) Example of the the field $f_{\text{SPSR}}(\mathbf{x})$. (b) Estimated uncertainty $\sigma(\mathbf{x})$. (c) Probabilistic query of occupancy $p\big(f_{\text{SPSR}}(\mathbf{x}) \le 0\big)$. (d) Gradient of $f_{\text{SPSR}}$ that the SPSR algorithm estimates from a simulated point cloud. (e) Surface reconstruction by running marching cubes with $f_{\text{SPSR}}=0$ and associated scalar variances on the grid (as visualised by the plasma colormap)}
    \label{fig:spsr_example}
\end{figure}

\subsection{Poisson Surface Reconstruction}

Poisson Surface Reconstruction (PSR)~\cite{kazhdan_poisson_2006} takes as input an oriented point set
$\mathcal{O} := \{ (\mathbf{p}_s, \mathbf{n}_s)\}_{s}$,
with surface samples $\mathbf{p}_s \in \mathbb{R}^3$ and associated normals $\mathbf{n}_s \in \mathbb{R}^3$, and constructs an implicit function, $f : \mathbb{R}^3 \rightarrow \mathbb{R}$.
This function takes positive and negative values outside and inside the surface respectively.
The zero level set $\{ \mathbf{x} \in \mathbb{R}^3 \mid f(\mathbf{x}) = 0 \}$ reconstructs the surface consistent with $\mathcal{O}$.
The function is obtained by solving a partial differential equation that recovers a scalar field whose gradient matches the interpolated normal field induced by the data~\cite{kazhdan_poisson_2006}.

\subsection{Stochastic Poisson Surface Reconstruction (SPSR)}
Stochastic Poisson Surface Reconstruction (SPSR)~\cite{sellan_stochastic_2022} extends PSR by placing a Gaussian process prior over the implicit function and performing Bayesian inference conditioned on $\mathcal{D}$.
The resulting posterior field satisfies
\begin{equation}
\label{eq:map_likelihood}
f(\mathbf{x}) \mid \mathcal{O} \sim \mathcal{N}\!\big(f_{\text{SPSR}}(\mathbf{x}), \sigma^2(\mathbf{x})\big),
\end{equation}
for all $\mathbf{x} \in \mathbb{R}^3$, and $\sigma^2(\mathbf{x})$ is the computed variance.

From an implementation perspective, the posterior field $f_{\text{SPSR}}$ is implemented via a discretized grid of resolution $K \times K \times K$, and the implicit, continuous model  is obtained via trilinear interpolation. However, in the remainder of the paper we use $f_{\text{SPSR}}$ indistinctively for simplicity.

The SPSR method also takes as input an oriented point set $\mathcal{O}$ and its noise parameters $\sigma_{\mathcal{O}}$, and returns the field $f_{\text{SPSR}}(\mathbf{x})$ and its variance $\sigma(\mathbf{x})$ corresponding to the  scalar implicit function $f$ evaluated on $\mathbf{x}$.
\begin{equation}\label{eq:spsr_operator}
    f_{\text{SPSR}}(\mathbf{x}), \sigma(\mathbf{x}) = \text{SPSR}(\mathcal{O}, \sigma_{\mathcal{O}})
\end{equation}
Optionally, SPSR also enables to initialize the method with a prior scalar field $f_{\text{prior}}(\mathbf{x})$.
\begin{equation}
    \label{eq:spsr_operator}
    f_{\text{SPSR}}(\mathbf{x}), \sigma(\mathbf{x}) = \text{SPSR}(\mathcal{O}, \sigma_{\mathcal{O}}, f_{\text{prior}}(\mathbf{x}))
\end{equation}
While \citet{sellan_stochastic_2022} use it to specify shape priors (spherical, cylindrical priors for objects), ContactFusion exploits this property to implement an online sequential estimation procedure.

%% file: content/method_revised_v2.tex
\section{Method}
\begin{figure*}[t]
\centering
\includegraphics[width=\linewidth]{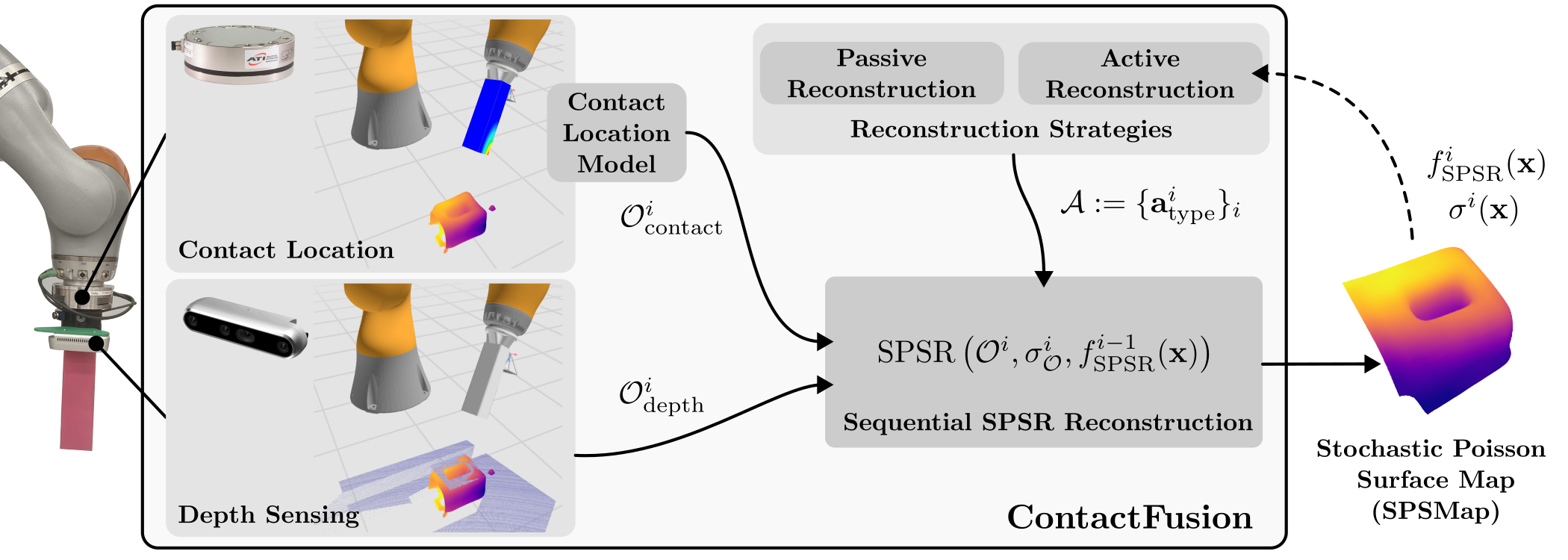}
\caption{\textbf{ContactFusion system overview.} Depth measurements are fused with Force-Torque derived spatial likelihoods into an uncertainty aware geometric representation. We propose using the Stochasitic Poisson Surface Reconstruction \cite{sellan_stochastic_2022} algorithm to sequentially fuse depth and F/T derived contact locations into a probabilistic map.}
\label{fig:system_overview}
\end{figure*}

\figref{fig:system_overview} shows the main steps of the ContactFusion system.
The inputs to our system are depth measurements from an RGB-D sensor, as well as force-torque measurements from a sensor located at the robot's wrist. Its output is a Stochastic Surface Reconstruction Map (SPSMap), a field that implicitly encodes the geometry and uncertainty of the map.

\subsection{Probabilistic Surface Mapping}
\label{subsec:mapping}
ContactFusion implements a sequential mapping procedure by integrating visual and contact measurements in the form of oriented point sets $\mathcal{O}$. By relying on the SPSR method, it implements online reconstruction by exploiting the operator defined by \eqref{eq:spsr_operator}. This motivates the following sequential update rule:
\begin{equation}
    \label{eq:spsr_update}
    f_{\text{SPSR}}^{i}(\mathbf{x}), \sigma^{i}(\mathbf{x}) = \text{SPSR}\left( \mathcal{O}^{i}, \sigma_{\mathcal{O}}^{i}, f_{\text{SPSR}}^{i-1}(\mathbf{x}) \right),
\end{equation}
where $i$ indexes the measurements, and $f_{\text{SPSR}}^{i-1}$ is the estimated scalar field obtained in the last iteration.

This mapping procedure is implicit, since SPSR defines the map via the field $f_{\text{SPSR}}$. Therefore, it does not provide direct access to occupancy or the surface itself. However, since it defines a probability distribution of the map, it enables closed-form geometric queries. 

The occupancy probability at a query point $\textbf{x}$ is defined by the cumulative distribution up to the zero-level set:
\[
p\big(f_{\text{SPSR}}(\mathbf{x}) \le 0\big)
=
\mathrm{CDF}_{f_{\text{SPSR}}(\mathbf{x})}(0),
\]
and the probability that $\mathbf{x}$ lies on the surface is given by the zero-level set:
\[
p\big(f_{\text{SPSR}}(\mathbf{x}) = 0\big)
=
\mathrm{PDF}_{f_{\text{SPSR}}(\mathbf{x})}(0).
\]

\figref{fig:spsr_example} illustrates the field that induces the surface as the zero level set (a), its uncertainty (b) and the estimated occupancy (c).
We note that the SPSR method yields smooth surface estimates while retaining a probabilistic interpretation of occupancy through the closed-form queries of the implicit function defined above.
This makes it well-suited to settings in which local measurements (such as contact) must refine a globally-coherent surface estimate from other modalities, like vision.
The next sections provide details about the measurements used for each modality.

\subsection{Vision-based Measurements}
\label{sec:visual_measurement}
A depth observation $\mathcal{O}_{\text{depth}}^{i}$ is obtained from a depth sensor measurement (e.g, RGB-D camera) and converted into an oriented point set:
\begin{align}
\mathcal{O}_{\text{depth}}^{i} := \{ (\mathbf{p}_s, \mathbf{n}_s) \}_{s},
\end{align}
where $\mathbf{p}_s \in \mathbb{R}^3$ are 3D points extracted from the depth measurement. $\mathbf{n}_s \in \mathbb{R}^3$ are corresponding surface normals estimated from local geometry, using the K-Nearest Neighbours method implemented in Open3D~\cite{zhou_open3d_2018}. We assume a fixed isotropic variance given by $\sigma^i_\text{depth}$.

\subsection{Contact Location Measurements}
\label{sec:contact_measurement}
While the depth sensor directly outputs the depth observations, a force-torque measurement does not directly yield a contact point. Instead, it constrains the set of possible contact locations consistent with the measured wrench. For this, we introduce a \textit{contact location model} to map force-torque measurement to an oriented point set of contact hypotheses.

Given a force-torque measurement $\mathcal{F}^i = (\mathbf{f}_{\text{obs}}^i, \boldsymbol{\tau}_{\text{obs}}^i)$ expressed in the sensor frame, the estimator computes candidate contact locations on the peg surface that are consistent with rigid-body mechanics.
Let $\mathbf{p}_{\text{can}}$ denote a candidate point on the peg surface and $\mathbf{p}_\text{origin}$ a point at the origin of the sensor frame.
Under a single-point contact assumption, the measured torque at the contact point should satisfy
\begin{align}
\boldsymbol{\tau}_{\text{can}}^{i} \approx (\mathbf{p}_{\text{can}} - \mathbf{p}_{\text{origin}}) \times \mathbf{f}_{\text{obs}}^i.
\end{align}
We therefore define a residual function
\begin{align}
l(\mathbf{p}_{\text{can}}; \mathcal{F}^i) = \left\| (\mathbf{p}_{\text{can}} - \mathbf{p}_{\text{origin}}) \times \mathbf{f}_{\text{obs}}^i - \boldsymbol{\tau}_{\text{obs}}^{i} \right\|^2,
\end{align}
which measures the consistency between the candidate contact location $\mathbf{p}_{\text{can}}$ and the measured wrench $\boldsymbol{\tau}_{\text{obs}}$.

\begin{figure*}[t]
    \centering
    \includegraphics[width=\textwidth]{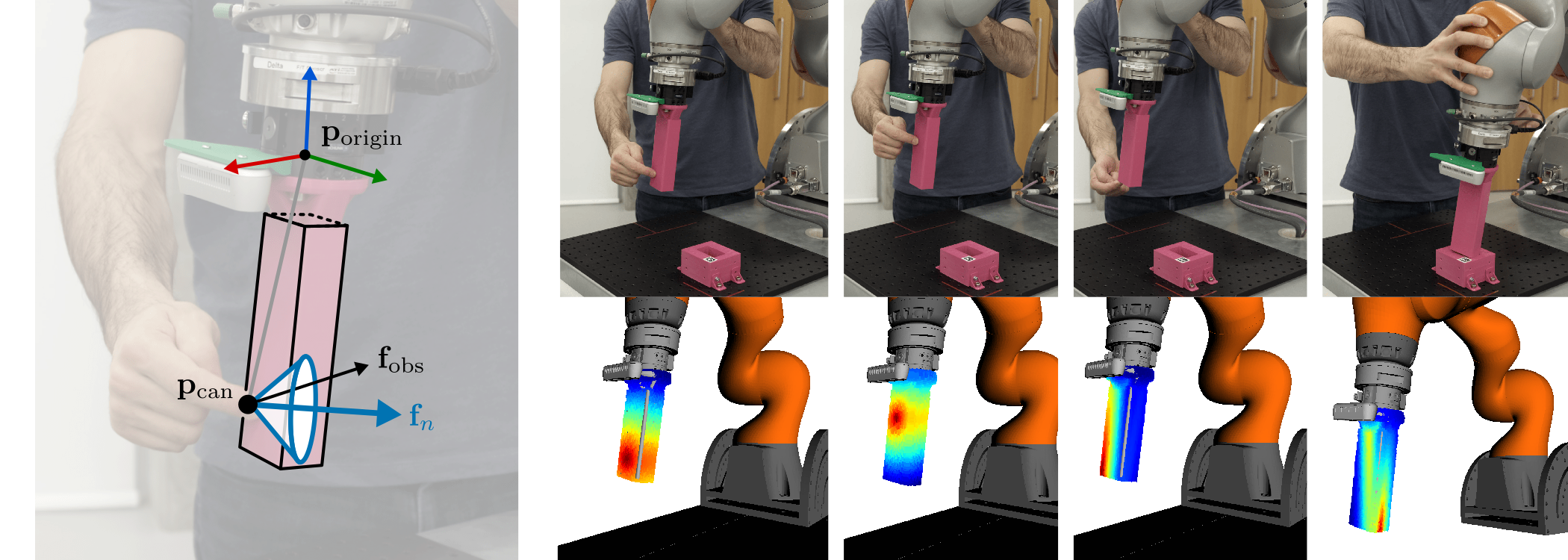}
    \caption{
        \textbf{Proposed contact location estimator.} Left: Force-torque residual model. $\mathbf{p}_{O}$ represents sensor origin, $\mathbf{p}_{C}$ represents sample where likelihood is being computed, $\mathbf{f}_{O}$ represents translational element of the wrench. Right: Examples of the contact location sensor from real interactions, where the color denotes the likelihood (blue: lowest, red: highest).
    }
    \label{fig:contact_locator_model}
\end{figure*}

To construct the contact observation $\mathcal{O}^i_{\text{contact}}$, we collect $K$ observations on the peg surface over a time-horizon, and retain those whose residual falls below a predefined threshold.
This procedure is equivalent to a rejection sampling approach: samples inconsistent with the measured wrench are rejected, while those satisfying the residual constraint are accepted as plausible contact hypotheses.
The associated contact normal for observation $k$ is taken to align with the measured normal force direction $\mathbf{f}_{n}$,
\begin{align}
\textbf{n}^{k} = \frac{\mathbf{f}^{k}_{n}}{\|\mathbf{f}^{k}_{n}\|}.
\end{align}
The resulting set of accepted candidates forms the oriented point set:
\begin{align}
\mathcal{O}^i_{\text{contact}} := \{ (\mathbf{p}_{\text{can}}^{k}, \mathbf{n}^{k}) \}_{k \in K}.
\end{align}
Optionally, a friction cone constraint may be imposed to further eliminate geometrically infeasible contact hypotheses. Similarly to the depth observations, we also consider a fixed isotropic variance $\sigma^i_\text{contact}$.
The complete procedure and real examples of the contact location model are shown in \figref{fig:contact_locator_model}.

\subsection{Visuo-Tactile Reconstruction Strategies}
\label{subsec:active_reconstruction}
In contrast to vision or LiDAR-based reconstruction approaches, a visuo-tactile system needs to explicitly establish contact with the environment to acquire new measurements. ContactFusion implements two reconstruction strategies, which determine how vision and contact observations are obtained and integrated. Specifically, they determine a sequence of actions indexed by $i$: 
\begin{align}
\mathcal{A}:= \{ \mathbf{a}^i_{\text{type}}\}_{i},
\end{align}
where $\mathbf{a}^i_{\text{type}}$ can be vision action  $\mathbf{a}^i_{\text{vision}}$ or a probing action $\mathbf{a}^i_{\text{probing}}$. We assume that each action is followed by an observation. These are illustrated in \figref{fig:action_space}.

A visual action specifies a camera viewpoint relative to a planar center in the end-effector frame,
\begin{align}
\textbf{a}^i_{\text{vision}} := (x_v,\, y_v,\, \alpha,\, \beta),
\end{align}
where $(x_v, y_v)$ defines the center of the viewing configuration in the $xy$-plane, and $\alpha$ and $\beta$ represent elevation and azimuth angles defining a single keyframe viewpoint. This defines a hemisphere of rays the robot can align to.

A probing action specifies a planar target location and a commanded translation along the negative $z$-axis of the end-effector frame,
\begin{align}
\textbf{a}^i_{\text{probing}} := (x_p,\, y_p,\, z_p),
\end{align}
where $(x_i, y_i)$ parameterizes the probe location in the $xy$-plane and $z_p$ is the magnitude of the translation along $-z$. This defines a 2D grid of probing locations.

These actions are executed using linear interpolated motion plans tracked via a compliant impedance controller.

\paragraph{Passive Reconstruction}
\label{par:passive_reconstruction}
The first strategy uses a pre-determined sequence of actions for the mapping process. We assume a fixed budget of $N$ measurements for this process. The first $\gamma$ correspond to vision actions, while the remaining $N-\gamma$ correspond to probing actions. Both vision and probing actions are sampled randomly, given some pre-defined limits to the parameters that define each space.

\paragraph{Active Reconstruction}
\label{par:active_reconstruction}
The second strategy exploits the uncertainty estimate of the surface to guide the exploration, effectively implementing a \textit{next best action} approach. Similarly to the passive strategy, we assign a budget of $N$ measurements for the experiment with $\gamma$ vision actions. However, instead of sampling, we use a greedy strategy guided by estimated uncertainty of the map at $i$.

For vision actions, we choose the ray in the vision action space that intersects the area of the mesh with higher uncertainty. Similarly, for probing actions we project the uncertainty to the 2D grid and vertically probe the map point with higher value.

%% file: content/evaluation_revised.tex
\section{Experimental Setup}
We design three experiments to validate the advantages of the probabilistic maps produced by ContactFusion in different tasks, such as geometry estimation, and pose estimation of the insertion target.
\begin{figure}[t]
    \centering
    \includegraphics[width=1\linewidth]{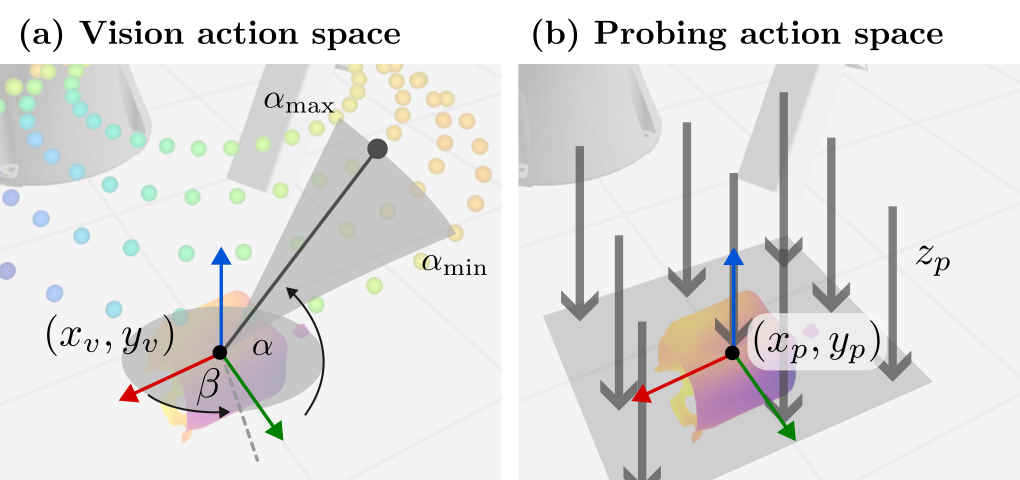}
    \caption{\textbf{Action spaces for the reconstruction strategies.} (a) The \textit{vision action space} considers a set of viewing angles centered at the current reconstruction. (b) The \textit{probing action space} defines vertical actions w.r.t a plane defined on the upper face of the current map. Next-Best Action scores are computed by projecting vertex uncertainty to the vision action points. For contact, high uncertainty vertices are chosen for vertical probing.}
    \label{fig:action_space}
\end{figure}

\begin{figure*}[h]
    \centering
    \includegraphics[width=1\linewidth]{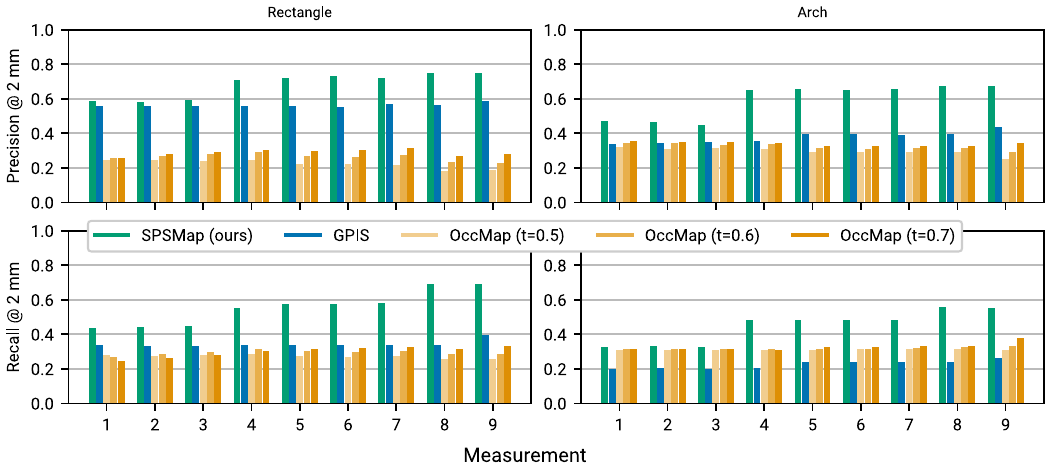}
    \caption{\textbf{Exp. 1: Insertion Target Reconstruction.} We report the Precision and Recall metrics averaged over 10 sample scenarios for the \textsc{Arch} and \textsc{Rectangle} geometry at a $\tau = 2 \text{mm}$. Large precision indicates a high proportion of the reconstructed mesh is under this $\tau$. Large recall indicates a high proportion of the ground truth mesh under $\tau$. Measurements 1-3 are depth based measurements acquired from pre-scripted robot panning motion. Measurements 3-9 are contact location measurements acquired from pre-scripted robot probe motions. We report surface reconstruction for OccMap at 3 occupancy thresholds, t = $\{0.5, 0.6, 0.7\}$}
    \label{fig:spsrvsomvsgpis}
\end{figure*}

\subsection{Exp. 1: Insertion Target Reconstruction}
The first experiment validates the reconstruction quality of insertion targets compared to other baselines. We compare the SPSMap, generated via \textit{ContactFusion}, against other mapping algorithms used in manipulation settings: Gaussian Process Implicit Surface (GPIS)~\cite{caccamo_active_2016, suresh_tactile_2021}, and Occupancy Maps (OccMap)~\cite{murali_shared_2024, marques_map_2025, dengler_viewpoint_2023}.

For SPSMap, we use a discretized grid at a resolution of 5mm.
We specify noise over depth and contact normals, $\sigma_\text{depth}^i = 0.1$ and $\sigma_\text{contact}^i=0.05$ respectively.
To compute the zero-level set, we use the marching cubes algorithm to recover the surface reconstruction, with the corresponding mesh vertex variances being queried from the grid.

For Occupancy Maps we use the same discretization as SPSMap---a voxel grid with a resolution of 5mm.
As in \citet{murali_shared_2024}, we uses a log-odds update rule, assigning a hit probability to be 0.7 and miss probability at 0.4.
To reconstruct the map, a threshold value of 0.5 is used to run the marching-cubes algorithm.

For GPIS, we use an enclosing grid of the same dimension as used for SPSMap and OccMap with an additional bound parameter.
This is used to specify exterior points to the point set as this is required to train the GPIS. We use the Thin Plate Spline kernel, as it was able to extrapolate asymmetric geometry better than the Radial Basis Function kernel. We also specify the centroid of the point set as the internal point.
We use the same $\sigma_\text{contact}^i$ and $\sigma_\text{depth}^i$ as used for SPSMap. 

This experiment is performed in simulation using Drake~\cite{tedrake_drake_2019} 
We record fixed sequences of the arm using the passive reconstruction strategy: we allocated a budget of $N=9$ observations with $\gamma=3$, i.e., 3 vision actions followed by 6 probing actions. 


\subsection{Exp. 2: Active Reconstruction}
The second experiment validates the advantages of the uncertainty estimates provided by SPSMap to guide an active reconstruction task, using the active strategy presented in \secref{subsec:active_reconstruction}.
Similarly to Exp. 1, we execute this in simulation with the same robot model and sensors. 
We use an occupancy map-based frontier exploration using information gain strategies \citet{murali_shared_2024} as the main baseline.

For our proposed active reconstruction strategy, we set $N=5$ and $\gamma=2$---2 vision actions and 3 probing actions. For the vision action, we partition the elevation parameter, $\alpha$ between $[\SI{20}{\degree}, \SI{40}{\degree}]$ at a resolution of $\SI{5}{\degree}$, and the azimuth parameter $\beta$ between $[\SI{0}{\degree}, \SI{360}{\degree}]$ at a resolution of $\SI{11.25}{\degree}$. For the probing action grid we used a resolution of 5mm. 

We randomize the hole position and collect 10 sequences of active mapping trajectories. The geometry reconstructions produced by both methods are then compared with the ground truth mesh. 

\subsection{Exp. 3: Real Robot Validation}
This last experiment validates ContactFusion in real robot settings.
We test on a KUKA IIWA arm, equipped with a Realsense D435 RGB-D camera, and an ATI force/torque sensor, both wrist mounted.
The ground truth location of the insertion target w.r.t the robot was obtained through extrinsic calibration. 
We used a passive reconstruction strategy with pre-defined motions to collect visual and contact information.

We randomize the target poses across 5 separate sequences. We compare the reconstructions with ground truth calibrated meshes.

\subsection{Metrics}
Given we are evaluating insertion tasks, we are concerned about evaluating the accuracy of the reconstructed target. Therefore, we use the precision (Pr), recall (Re) and F-Score metrics setting a desired tolerance $\tau$.

To determine these metrics, we compute point-to-point error distances between transformed ground truth (GT) mesh and the geometry prediction $\mathcal{M}$, following the methodology of \citet{seitz_evaluation_2006}.
To do this we sample $L=10,000$ points on both meshes and compute two-way distance between $e_{G} = \text{GT} \rightarrow \mathcal{M}$ and $e_{\mathcal{M}} = \mathcal{M} \rightarrow \text{GT}$.
The precision metric reports the \textit{accuracy} of the reconstruction at a desired tolerance, where higher precision means a closer fit to the ground truth.
The recall metric reports the proportion of GT points that are below an error threshold, $\tau$, representing the \textit{completeness} of the map for the given tolerance. Higher recall means a more complete coverage of the ground truth

Combining these metrics we also report the F-Score:
\begin{equation}
    \text{F} = \frac{2\text{PR} \cdot\text{Re}}{\text{PR}+\text{Re}}.
\end{equation}

%% file: content/results.tex
\section{Results}
\subsection{Exp. 1: Insertion Target Reconstruction}
For the simulation experiment, \cref{fig:spsrvsomvsgpis} reports precision and recall metrics averaged over 10 sample scenario at a error threshold $\tau = 2\text{mm}$ across 9 measurements.
Our experimental findings indicate that SPSMap and GPIS recover higher precision and recall when compared to OccMap.
We find the SPSMap and GPIS representations extrapolate to unobserved geometry as is evident with the higher recall metrics across both the Rectangle and Arch geometries.
Contact measurements improve local geometric conditioning, as can be observed in the increased trends within both precision and recall after measurement 4 on both geometries.
Both SPSMap and GPIS provide alternative belief representations over geometry for the insertion task that are able to integrate contact and depth information.

\begin{figure}[t]
    \centering
    \includegraphics[width=1\linewidth]{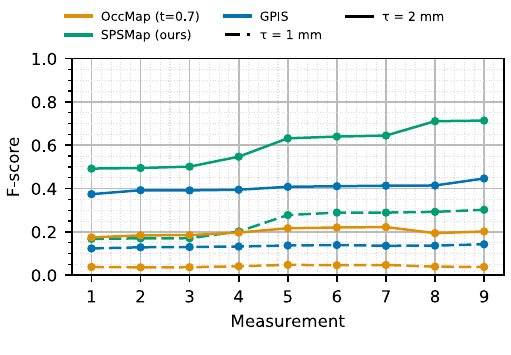}
    \caption{\textbf{Exp. 1: Avg F-scores at different tolerances.} We report the F-score averaged across 10 scenarios, as a function of discrete depth and contact measurements. Measurements 1-3 are depth measurements, measurements 4-9 are contact probes.}
    \label{fig:avg_f_scores}
\end{figure}

Performance of the representations at $\tau = \{1\text{mm}, 2\text{mm}\}$ across 10 scenarios are summarized by their F-Scores in \cref{fig:avg_f_scores}.
We report that SPSMap (in green) has a consistently higher F-Score across measurements, of up to 30\%.

\subsection{Exp. 2: Active Reconstruction}
\begin{figure}
    \centering
    \includegraphics[width=\linewidth]{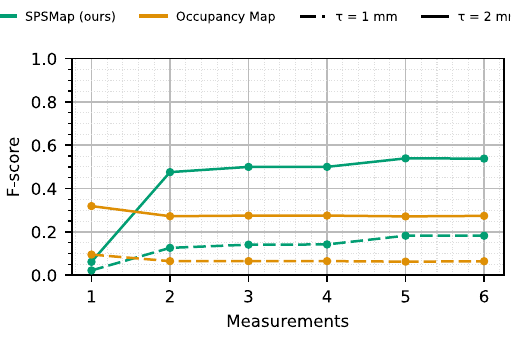}
    \caption{\textbf{Exp. 2: Active reconstruction}. Uncertainty after each measurement is used to decide Next-Beset Action. Measurements 1-2 are via depth, measurements 3-6 are contact probes.}
    \label{fig:active_recon}
\end{figure}

Finally, \cref{fig:active_recon} reports our F-Scores on the active reconstruction experiments.
SPSMap consistently outperforms the occupancy map baseline in reconstruction quality. 
This is evident within the increasing trend of the F-Score across 6 measurements averaged across 6 scenarios.
The surfaces reconstructed using SPSMap are smoother and more geometrically consistent, particularly in sparsely observed regions.
In contrast, occupancy maps often produce fragmented reconstructions due to their locally independent grid updates.

\subsection{Exp. 3: Real Robot Validation}
\begin{figure}[t]
    \centering
    \includegraphics[width=1\linewidth]{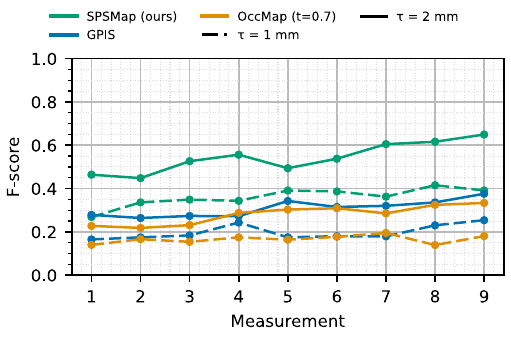}
    \caption{\textbf{Exp. 3: Real Robot Validation.} Avg F-scores at different thresholds, $\tau$, averaged across 5 real robot scenarios. Measurements 1-3 are via depth, measurements 4-9 are contact probes.}
    \label{fig:real_robot}
\end{figure}

For the real robot experiment, \cref{fig:real_robot} shows the average F-Score for the 5 sequences collected over randomized hole positions.
We report that SPSMap has a higher F-Score of up to 35\%.
Furthermore, for both Exp. 1 and Exp. 2 we observe that SPSMap consistently recovers better precision and recall on symmetric and assymetric objects over GPIS.
We hypothesise that the improved performance of SPSMap over GPIS arises from the inductive bias imposed by the positive definite kernels used in GPIS. These kernels enforce symmetric covariance structures in the implicit function, which can bias extrapolation in sparsely observed regions, whereas the SPSR-based formulation underlying SPSMap imposes weaker symmetry assumptions and can therefore better capture asymmetric or partially observed geometries.

\begin{figure}[t]
    \centering
    \includegraphics[width=0.95\linewidth]{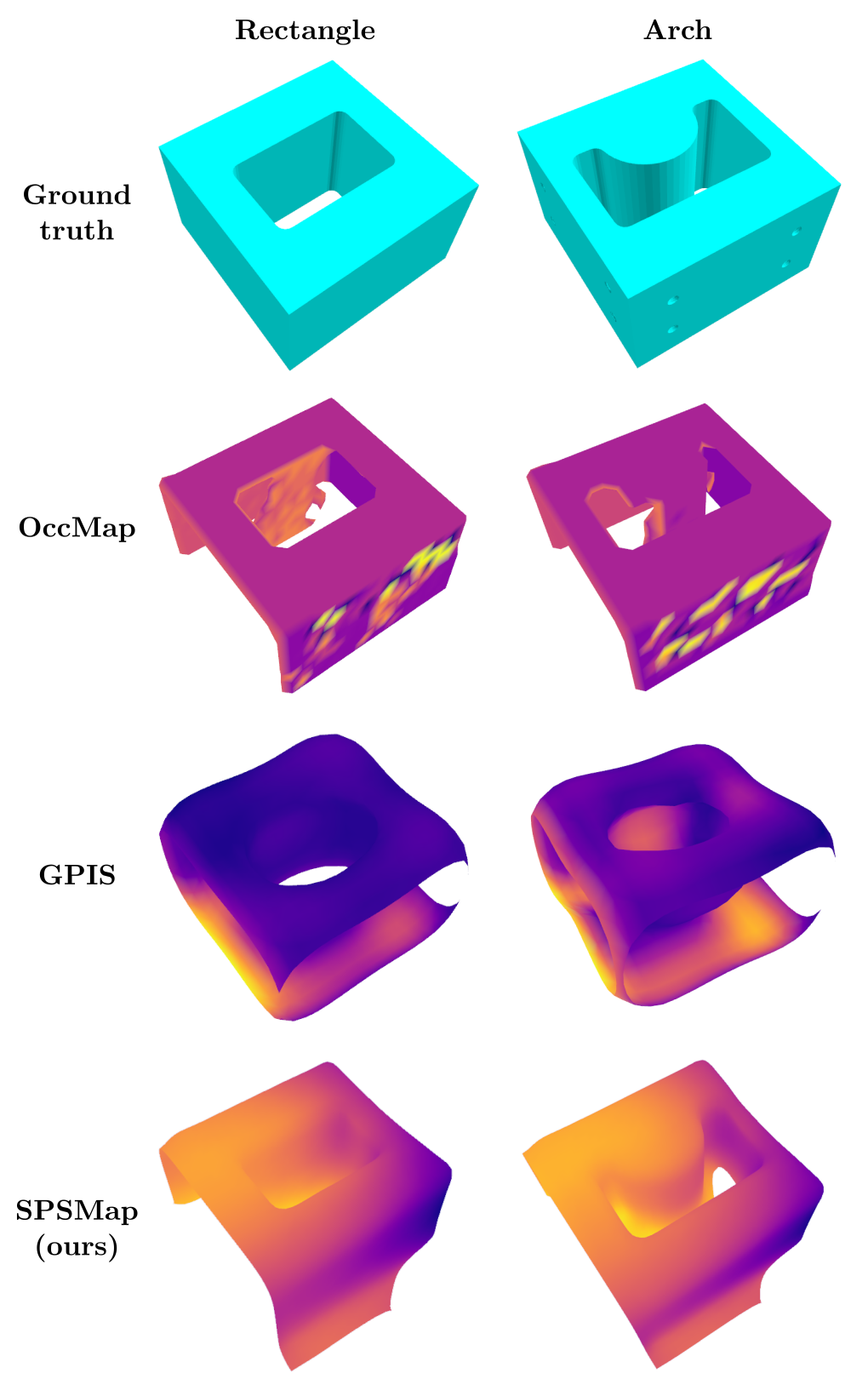}
    \caption{\textbf{Exp. 1: Qualitative results.} We show examples of the different baselines methods for the arch and rectangle geometries.}
    \label{fig:qualitative_results}
\end{figure}

\subsection{Computation Time}
We note that the benefits of SPSMap come at a significant time complexity.
\cref{tab:reconstruction_comparison} shows time taken to update each representation with a measurement. 
SPSMap involves solving a Partial Differential Equation at measurement update.
The time taken is a function of the underlying discretization, and solving the associated linear system \cite{sellan_stochastic_2022}\cite{kazhdan_poisson_2006}.
In contrast, OccMap involves updating the full grid with hit and miss probabilities, which is made efficient with modern mapping libraries.
Despite this additional computational overhead, SPSMap consistently produces more accurate and geometrically coherent reconstructions, and future work could focus on accelerating the underlying FEM/PDE update by parallelizing the associated sparse linear system solve.

\begin{table}[t]
    \centering
    \caption{\textbf{Computation time.} We report the average F-Score and time complexity of reconstruction methods on real robot data. Time complexity is averaged over each measurement update. }
    \label{tab:reconstruction_comparison}
    \begin{tabular}{lcc}
    \toprule
    \textbf{Method} & \textbf{Avg. F-Score} $\uparrow$ & \textbf{Time (ms)} $\downarrow$ \\
    \midrule
    SPSMap (Ours) & 0.649 & 9,900 \\
    GPIS & 0.376 & 5,800 \\
    Occupancy Map & 0.334 & 200 \\
    \bottomrule
    \end{tabular}
\end{table}


%% file: content/conclusion.tex
\section{Conclusion}

In this work, we introduced \textit{ContactFusion}, a probabilistic mapping framework that fuses visual and contact sensing to estimate the geometry of insertion targets in manipulation tasks.
By leveraging Stochastic Poisson Surface Reconstruction (SPSR) \cite{sellan_stochastic_2022}, our approach constructs an uncertainty-aware implicit representation of the surface, SPSMap, that can be sequentially updated with new measurements.

We presented a contact location estimator that converts force–torque measurements into spatial hypotheses of contact locations, enabling integration with depth observations within the same probabilistic reconstruction framework.
Through simulation and real-robot experiments, we demonstrated that the resulting Stochastic Poisson Surface Map (SPSMap) provides more accurate and geometrically coherent reconstructions than GPIS and occupancy map baselines, achieving improvements of up to 30–35\% in reconstruction F-score.
Furthermore, the uncertainty estimates produced by SPSMap enable active sensing strategies that improve reconstruction efficiency.